\documentclass[conference]{IEEEtran}
\usepackage{cite}
\usepackage{float}

\ifCLASSINFOpdf
\else
\fi
\usepackage[caption=false,font=footnotesize]{subfig}
\usepackage{pgfplots}

\newcommand{\argmax}{\operatornamewithlimits{argmax}}

\ifCLASSINFOpdf
   \usepackage{graphicx}
\else
\fi

\usepackage[cmex10]{amsmath}
\usepackage{color, framed, tikz}
\usepackage{fancyhdr}

\renewcommand{\thispagestyle}[2]{}

\fancypagestyle{plain}{
        \fancyhead{}
        \fancyhead[C]{first page center header}
        \fancyfoot{}
        \fancyfoot[C]{first page center footer}
}
\begin{document}

%
\title{Automatic Extraction of Causal Relations from Natural Language Texts: A Comprehensive Survey}

\author{\IEEEauthorblockN{Nabiha Asghar}
\IEEEauthorblockA{David R. Cheriton School of Computer Science\\
University of Waterloo\\
Waterloo, ON, N2L 3G1\\
Email: nasghar@uwaterloo.ca}}


%


\maketitle

\begin{abstract}
Automatic extraction of cause-effect relationships from natural language texts is a challenging open problem in Artificial Intelligence. Most of the early attempts at its solution used 
manually constructed linguistic and syntactic rules on small and domain-specific data sets. However, with the advent of big data, the availability of affordable computing power and the recent popularization of machine learning, the paradigm to tackle this problem has slowly shifted. Machines are now expected to learn generic causal extraction rules from labelled data with minimal supervision, in a domain independent-manner. In this paper, we provide a comprehensive survey of causal relation extraction techniques from both paradigms, and analyse their relative strengths and weaknesses, with recommendations for future work.
\end{abstract}



%

\section{Introduction}
\label{intro}

Automated knowledge extraction from natural language texts is one of the many open challenges in Aritificial Intelligence today. 
Effectively mimicing the human brain's ability to understand written texts entails the development of an intricate model to deal with the interplay of syntax, semantics, a continually evolving vocabulary, and ambiguous constructs like figurative expressions, metaphors, rhetorics, sarcasm and slang. Indeed, even a simple task like identifying implicit negations in a sentence remains an unsolved problem. Nevertheless, the Computer Science and Computational Linguistics communities have made significant breakthroughs in the area over the last three decades, and continue to churn progressively efficient, robust and scalable techniques specifically tailored for textual information extraction tasks.

In recent years, automatic extraction of semantic relations has become increasingly important for applications related to question answering \cite{girju2003}, information retrieval \cite{khoo1998}, event prediction \cite{radinsky2012, silverstein2000}, generating future scenarios \cite{riaz2010, hashimoto2014} and decision processing \cite{ackerman2012}. These relations, such as part-whole, if-then, cause-effect and so on, encode crucial information about how different events and entities should be perceived in relation to each other. In particular, the cause-effect relation is thought to play a very important part in human cognition due to its ability to influence decision making \cite{chan2005}. For instance, a tool to automatically scour the plethora of textual content on the world wide web and extract meaningful causal relations could help us construct \textit{causal chains} to discover previously unknown relationships between entities/events. This knowledge augmentation could be supremely valuable in many domains, including medicine \cite{khoo2000}, biology \cite{sachs2005} and environmental sciences \cite{arauz2012}. 

Causality has been studied extensively in a wide range of disciplines, including Psychology \cite{wolff2003, wolff2007}, Linguistics \cite{talmy2000, hobbs2005}, Philosophy \cite{white1990} and Computer Science. One of the simplest ways to express cause-effect relations is through propositions of the form `$A$ causes $B$' or `$A$ is caused by $B$'. It is a highly intuitive notion and yet, the topic has been surrounded by much controvery because experts belonging to these fields often disagree about when two events are causally linked. This is understandable, because causality can be expressed using many different types of propositions (e.g., active, passive, subject-object, nominal or verbal) and have several diverse syntactic representations. One popular classification of its explicit representations was given by Khoo \textit{et al}. \cite{khoo1998}.
\begin{enumerate}
\item \textit{Causal links} can be used to link clauses or sentences. Altenberg \cite{altenberg1984} classified causal links into four types: 1) adverbial links, e.g. \textit{so, hence, therefore}, 2) prepositional links, e.g. \textit{because of, on account of}, 3) subordination, e.g. \textit{because, as, since}, and 4) clause-integrated links, e.g. \textit{that's why, the result was}.
\item \textit{Causative verbs} are transitive verbs whose meanings include a causal element \cite{thomson1987}. Examples include \textit{break} and \textit{kill}, whose transitive forms are \textit{to cause to break} and \textit{to cause to die}.
\item \textit{Resultative constructions} are sentences in which the object of a verb is followed by a phrase describing the state of the object as a result of the action denoted by the verb. Examples include `I painted the car \textit{yellow}' and `I cooked the meat to a \textit{cinder}' \cite{simpson1983}.
\item \textit{If-Then conditionals} often indicate that the antecedent causes the consequent. 
\item \textit{Causation adverbs and adjectives} have causal element in their meanings, e.g. \textit{fatal} or \textit{fatally}, that can be paraphrased as \textit{to cause to die} \cite{cresswell1981}.
\end{enumerate}

It is clear that the syntax of each of these representations can vary, and these variations are difficult to capture and formalize using a single grammatical model. In fact, other studies propose entirely different classifications of syntactical representations of causal relations (see \cite{nastase2004, blanco2008}), each with its own merits and de-merits. Moreover, these classifications are not generalizable to non-English languages such as Chinese, Arabic \cite{sadek2013} and French \cite{garcia1997}. Therefore, in spite of its significance and applicability to a wide array of domains, automatic causality extraction remains a hard NLP problem to solve.

The existing literature on causal relation extraction falls into two broad categories: 1) approaches that employ linguistic, syntactic and semantic pattern matching only, and 2) techniques based on statistical methods and machine learning. In this paper, we survey some of the most promising causal relation extraction techniques proposed in the last three decades. Sections \ref{nonmlbased} and \ref{mlbased} provide a literature review and discussion on research studies belonging to each of the two categories respectively. In section \ref{analysis}, we analyse their relative strengths and weaknesses.  We conclude in section \ref{conc} with propositions for future work.

\section{Non-Statistical Techniques}
\label{nonmlbased}

The idea of automatic information extraction from natural language texts took off in the 80s, in the wake of increasingly affordable computational power and the rising popularity of machine learning.  In 1989, Selfridge \cite{selfridge1989} wrote an interesting paper on why it would be very hard to develop an automatic natural language-based causal knowledge acquisition system. The author highlighted that a good causal analyser must necessarily arise from a knowledge representation model that incorporates  a semantic framework, and it must simultaneously possess domain knowledge in order to accurately identify \textit{meaningful} causal relations from any text. Additionally, it was asserted that the tool should be able to infer any unmentioned immediate components between the causal component and the effect component in order to produce a more accurate and comprehensive  \textit{causal chain}, and should be able to disambiguate any causally ambiguous statements. 

\subsection{The Early Years: Small Texts, Manual Annotation, Hand-coded Features and Domain Dependence}
Several research studies in the late 80s and 90s attempted to tackle the problems highlighted by Selfridge. One of the first prominent works that developed a fully functional automated causal relations extractor from expository\footnote{This is the kind of text found in encyclopedias and textbooks.} English text was by Kaplan and Berry-Rogghe in 1991 \cite{kaplan1991}. Their tool starts with a text encoded into a set of propositions:

\vspace{-1.5mm}
\small
\noindent \textit{\underline{Sentence:}} When a cloud forms the water vapour condenses into water.
\noindent \textit{\underline{Proposition:}} when(event(form(result(cloud))), event(condense(patient\\(water vapour), state(water))).
\vspace{-1.5mm}
\normalsize


\noindent These propositions consist of a predicate, usually a verb, followed by one or more arguments that are marked with their syntactic roles. To enable generalization to complex sentences, propositions are allowed to take other propositions as arguments. These propositions are passed to a semantic analyser to produce a series of concept frames based on a knowledge representation 
In this representation, each `object', `action', and `relationship' is a concept with a pre-defined template. While parsing the text, the semantic analyser looks for these concepts, binds each occurence to its appropriate template and passes them to a causal analyser which searches for causally related concepts via 4 methods:
\begin{enumerate}
\item using 20 hand-coded explicit propositional clues designating causality, e.g. \textit{because, due to} and \textit{when};
\item extending the explicit cause-effect pairs found in 1) to causal chains. This is done by initializing with a \textit{seed} causal relation $X \rightarrow Y$, searching for a causal pair $A \rightarrow B$ such that $B=X$, and a causal pair $C \rightarrow D$ such that $Y=C$. This produces the chain $A \rightarrow X \rightarrow Y \rightarrow D$. The process can be repeated to further elongate the chain;
\item using Doyle's model of causality \cite{doyle1984} to find cause-effect pairs by finding pairs that are both temporally and spatially related; 
\item manually constructing constraints for several domain-specific concepts that may imply causality, e.g.  `increase(heightof(hot-air)) $\equiv$ formation(cloud)'. Thus, implicit causalities such as `\textit{warm air rises and we see clouds}' are captured by these constraints.
\end{enumerate}
  
There are several limitations to this knowledge-based inference approach. It requires extensive manual pre-processing, such as encoding the text into propositions, constructing propositional clues to detect explicit causal relations and designing constraints to detect implicit causalities. Moreover, these steps are pre-dominantly domain specific; expert knowledge is required to design a handful of clues and constraints manually. In addition, it is not clear how the temporally and spatially related concepts are extracted in step 3. The data set is tiny, comprising at most a hundred expository sentences. Thus, narrative and episodic texts are ignored, as are ambiguous syntactic and semantic constructs like analogies, references and ellipses.
  

PROTEUS \cite{grishman1990} was a similar work that used syntactic and semantic analysis 
to produce a network of temporal and causal relations from a given text. However, the data set was small and domain specific, and the process involved extensive manual work.  Kontos and Sidiropoulou \cite{kontos1991} also dealt with expository text and used linguistic patterns to detect causal relations, but relied heavily on hand-coded patterns to handle the small and domain-specific texts. Garcia \cite{garcia1997} produced one of the next notable works in the era. Their tool, COATIS, exclusively searched for linguistic patterns containing 23 causal verbs to automatically extract causal relations from French texts. Special attention was paid to syntactic positions of these verbs and the surrounding noun phrases. Basic attempts were also made to handle frequently occurring ambiguities in French causal phrases. These factors enabled COATIS to model generic notions expressed by French language, and thus operate in a domain-independent fashion. It achieved a reasonably good precision of 85\%, but on a relatively small text. 

Garcia's COATIS, though only a prototype, was an effective demonstration of how domain-independence could be achieved. Christopher Khoo developed this idea in a series of seminal works published in 1995 up to 2000 \cite{khoo1998, khoo1999, khoo2000}. The scope of these works was limited to the detection of \textit{explicitly} expressed causal relations only, but they bypassed domain-dependence by foregoing knowledge-based inference and relying entirely on linguistic clues comprising four types of causal links, 2082 causative verbs, resultative constructions of the form \textit{Verb-NounPhrase-Adjective} and \textit{if-then} conditionals. We defined these terms in Section \ref{intro} with a few examples. Interested readers can view a more comprehensive list for each of these constructs in Khoo's Ph.D. thesis \cite{khoo1995}. In \cite{khoo1998}, the POS-tagged phrases/sentences are matched with the ordered list of linguistic patterns to obtain partial or complete matches. The list of linguistic patterns is ordered so that specific patterns occur before more general patterns. For example, the pattern \textit{`[cause] and because of this, [effect]'} is more specific than the patterns \textit{`because [cause], [effect]'} and \textit{`[effect] because [cause]'}. Also, the second pattern is more specific than the third pattern. This means that the second and third pattern will match all the sentences that the first pattern matches with. In particular, all three will match the sentence \textit{`It was raining heavily and because of this, the car failed to brake in time'}, but only the third pattern will match the sentence \textit{`The car failed to brake in time because it was raining heavily'}.

As is evident, the computation model used in this research was very simple and could be used to detect causality within sentences as well between adjacent sentences. The algorithm was evaluated on Wall Street Journal articles spanning four months, containing a total of 1082 sentences, and was compared to the list of causal relations identified and agreed upon by two human judges. For causal links the recall\footnote{Here, recall is the proportion of the causal relations identified by both human judges that were also identified by the algorithm.} was 78\% and the precision\footnote{Precision is the proportion of causal relations identified by the algorithm that were also identified by the human judges.} was 42\%, whereas for causative verbs the recall was 61\% and precision was 19\%. 
The overwhelmingly poor precision results could be attributed to lexical ambiguities. For instance, the algorithm did not take into account the fact that the same syntactic constructions used to indicate cause-effect can also be used to indicate other types of relations. Several mistakes were made due to complex sentence structures, and a few omissions occured because some causative verbs (e.g. \textit{lift}) occured in the text in a noun form, a non-trivial check that the algorithm failed to consider. 

Overall, most of the limitations of Khoo's work could be attributed to the use of explicit causal indicators only and ignoring implicit causalities. Moreover, the absence of inferencing from a knowledge-base and the use of a domain independent data set\footnote{In this case, the data set was much larger than those used in previously referenced works, but consisted of short and straightforward sentences only.} led to many lexical ambiguities. A tweaked version of this algorithm, which required a much smaller number of linguistic patterns, was applied to textual databases containing articles from the field of medicine and psychology \cite{khoo1999, khoo2000}, with the motivation that new knowledge could be synthesized using causal knowledge existing in published literature \cite{finn1998}. The improved precision and recall performance in these studies was attributed to the algorithmic tweaks. However, this claim is debatable; the improvement could very well be due to the data's domain specificity.

\subsection{The New Millenium: Focus on Large and Domain Independent Texts, Automation and Scalability}

Evidently, the problem needed to be revisited from a more pragmatic perspective. Relying solely on manually hand-crafted linguistic patterns and manually encoded propositions was impractical. There was a need to focus on automation, scalability and reduction of dependence on textual domains and types. Moreover, the more difficult problem of extracting implicit causal relations needed to be addressed with rigor. 
\begin{figure}
\centering
\includegraphics[width=68mm, height=73mm]{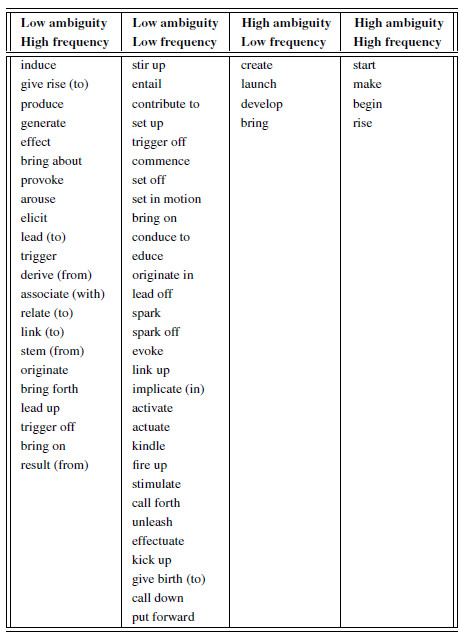}
\caption{Girju and Moldovan \cite{girju2002}'s ranking of ambiguous causation verbs, based on ambiguity and frequency.}
\label{girjumoldovan}
\end{figure}

A highly-cited work by Girju and Moldovan \cite{girju2002}, published in 2002, was a move towards this desirable research direction. They proposed an algorithm for automatic discovery of linguistic patterns expressing causal relations and a semi-supervised way to validate the list of obtained patterns based on semantic constraints on nouns and verbs. To simplify the problem, the authors focused on explicit syntactic patterns of the form \textit{NP1-CausativeVerb-NP2}\footnote{NP stands for noun phrase.}. A different linguistic classification of causative verbs was used in this paper \cite{nedjalkov1973}, whereby we could either have \textit{simple causatives} that were synonymous with the word `cause', \textit{resultative causatives} which were similar to the resultative constructions used by Khoo \cite{khoo1998}, and \textit{instrumental causatives} that express a part of the causing event as well as the result, e.g., `poison'. As a second simplification, the paper focused on simple causatives only. The algorithm used in this paper to extract lexico-syntactic patterns of the form \textit{NP1-SimpleCausativeVerb-NP2} is a variation of the famous Hearst's procedure \cite{hearst1998} for WordNet \cite{miller1995}. Concretely, the algorithm extracts all possible pairs of noun phrases between which WordNet causal relations\footnote{WordNet causal relations are extracted by looking at the gloss definition of each noun and extracting the nouns that appear as its causes or effects.} exist, and then for each pair \textit{NP1-NP2} found, it searches a collection of documents to find all the patterns of the form \textit{NP1-Verb/VerbExpression-NP2}. This procedure outputs a list of verbs or verb expression that refer to causation most of the times, but not always. For example, the verb \textit{produces} sometimes refers to a non-causal relationship between two objects. 
To deal with such ambiguities, each pattern is first validated through a WordNet-based course-grained process that detects the semantic constraints on the two nouns and the verb which necessarily and sufficiently indicate a causal relationship. In particular, for the \textit{effect} nouns, five causation classes are identified in WordNet by finding the most general subsumers of the \textit{effect} nouns: \textit{human action, phenomenon, state, psychological feature} and \textit{event}. Moreover, it is noted that the \textit{cause} nouns must have as a subsumer the concept \textit{causal agent} in WordNet. Semantic constraints on a verb are found by ranking the verb's \textit{senses}\footnote{In WordNet, there is a list of synonyms for each verb, called the verb's \textit{synset}. Thus, each verb can have multiple meanings, or \textit{senses}.} based on frequency and ambiguity. Finally, based on these constraints, the pattern is ranked on a scale of 1 to 4, with 1 denoting a highly likely causal relation and 4 denoting high ambiguity (see Figure \ref{girjumoldovan}).

Experiments were conducted on a large and diverse set of news articles to compare the algorithm's performance with the average of two sets of manual annotations. The algorithm's accuracy was 65.6\% and its ranking of the causal relationships matched the manual rankings by 52\%. While the numbers were not as high as those obtained by Khoo \cite{khoo1999, khoo2000} (due to domain independence), the proposed algorithm clearly had higher merit based on its novelty. This was the first algorithm to treat the problem of ambiguity in a hands-on, albeit semi-automatic, manner, and to generate the linguistic patterns automatically. 

The reliance of Girju and Moldovan's work on WordNet was an excellent attempt at handling implicit and ambiguous cases, and the ranking procedure was very useful and novel. However, there were some obvious limitations, in that they only considered explicit causal syntactic patterns of a particular form, and focused on extracting explicit and simple causative verbs only. Some of these problems were addressed by a group of researchers at the Chinese University of Hong Kong, in a series of works \cite{low2001, chan2002, chan2005} in the early 2000s. Their novel idea was that humans understand and extract knowledge from texts easily because they have some expected semantics for the text in mind, which guides their search and understanding of information in the text. For example, the expectations about the text while reading a newspaper are different from those while reading a story book or a novel. Therefore, expectation-based semantics should be used to build a generic and domain independent causal searching tool. To this end, they developed SEKE, a Semantic Expectation-based Knowledge Extraction framework for causal relations.

\begin{figure}
\centering
\includegraphics[width=70mm, height=73mm]{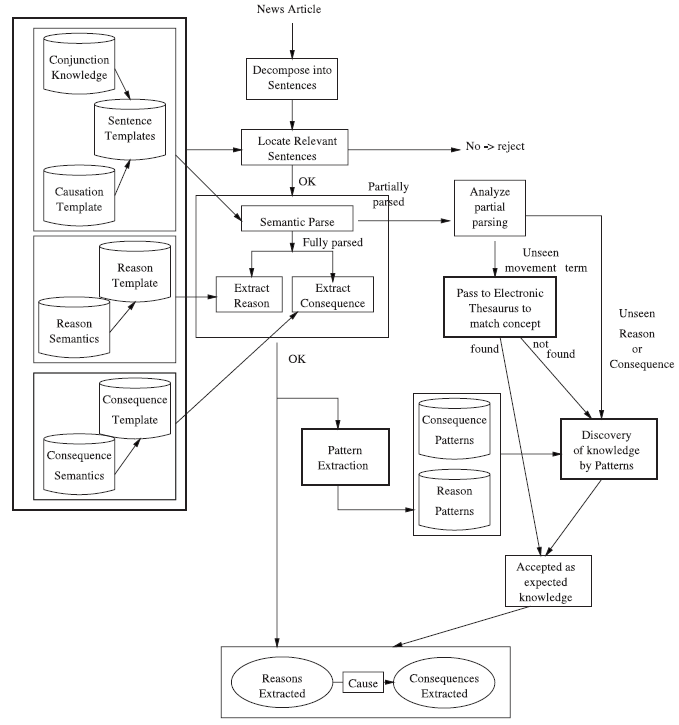}
\caption{The SEKE Architecture \cite{chan2005}.}
\label{seke}
\end{figure}

SEKE consists of several types of generic semantic  templates organized hierarchically. Their interplay can be seen in SEKE's architectural framework in Figure \ref{seke}.  As a pre-processing step, the templates are designed manually by picking out and semantically analyzing all the sentences containing causative expressions from a training corpus (in this case, 365 news articles about Hong Kong stock market). Only causal links are considered in this analysis. Thus, templates for simple and complex sentences are designed first, and then the templates for causes and effects are designed (called `reason and consequence templates' in Figure \ref{seke}). Once this seed knowledge representation is ready, SEKE processes the text by screening out all the non-causative sentences in the test data using the sentence template, and parsing the remaining sentences into causes and effects through pattern matching using the reasons and consequences templates. To account for complex sentences, the extracted causes and effects are recursively parsed to obtain deeper or implicit causal knowledge from them. If a phrase matches multiple patterns,  the patterns are ranked using a \textit{matching score}. This scoring function is carefully chosen to account for the support of the patterns as well as manual control parameters. Interested readers are referred to \cite{chan2005} for detailed explanations of the new pattern discovery and the scoring function, and several small illustrative examples. SEKE includes a WordNet-based component, used for finding concepts synonymous to the extracted case and effect phrases. This helps with disambiguation and also generates new causal patterns that are previously unseen. A separate pattern discovery component also resides in the SEKE framework, which takes the extracted patterns as input and generalizes them using concept labels and wildcards. 

The authors tested SEKE's performance on a test set of 365 Hong Kong stock market related news articles, and obtained a recall of 45.9\% and a precision of 71.6\%. More importantly, SEKE found several unseen patterns successfully. Thus, the authors successfully augmented Girju and Moldovan's idea of using WordNet and pattern ranking to handle ambiguities and to generate previously unknown knowledge. However, the authors did not test the system on domain independent data, so it is not clear how SEKE would scale to diverse texts.

The results of the techniques discussed so far established the existence of a trade-off between high accuracy (high precision and high recall) and scalability to multiple domains. Moreover, appropriately designed semantic ranking measures remained the most popular technique for handling ambiguous causal relations. Predominantly, no one paid much attention to the extraction of implicit causal patterns that have no specific linguistic form, i.e. they are not synonymous with \textit{to cause} but still influence the reader to causally link two or more events in the text. More recently, Ittoo and Bouma \cite{ittoo2011, ittoo2013} attempted to tackle this hard problem while reverting to domain specific texts as a simplification measure. They focused on three types of implicit causal relations: 
\begin{enumerate}
\item resultative and instrumentative verbal patterns e.g. \textit{increase, kill} which specify the resulting situation;
\item patterns that make the causal agents inseparable from the resulting situations, e.g. \textit{white spots mar the X-ray image};
\item non-verb patterns, e.g. \textit{due to}.
\end{enumerate}

\begin{figure*}
\centering
\includegraphics[width=157mm, height=40mm]{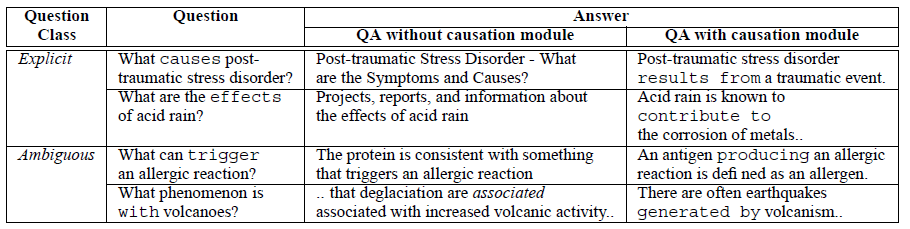}
\caption{Examples of cause-effect questions tested on a Question Answering system \cite{girju2003}.}
\label{girju03}
\end{figure*}

In addition, they provided a way to disambiguate polysemous\footnote{Polysemous means `having multiple meanings'. For example, the phrase \textit{lead to} can be used in causal as well as non-causal sense. Compare the sentence \textit{smoking leads to cancer} with the sentence \textit{paths lead to the garden}.} patterns.
 Concretely, their technique uses Wikipedia to build a knowledge base since it has broad coverage and is highly likely to contain a wide array of explicit and implicit linguistic patterns encoding causality. First, all the shortest patterns linking two noun phrases are extracted. To extract causal patterns from this list of generic patterns, the proposed algorithm starts with a seed causal pair (e.g. hiv - aids) and identifies patterns in Wikipedia that connect the seed pair. A reliability measure for each pattern is computed, the top $k$ most reliable patterns are compiled and then other pairs connected by these patterns are identified recursively. These steps are repeated iteratively for as many seed pairs as desired. This procedure of learning new patterns from pairs and vice versa is carried out till a suitable number of causal relations has been extracted. To evaluate the algorithm, the authors initialized it with a domain specific Wikipedia dump consisting of $\sim$ 500 million words. From the 2 million generic patterns extracted, the algorithm used 20 seed pairs to identify 81 causal patterns. On a test set of 32,545 domain specific documents, a precision of 0.765 and a recall of 0.82 was obtained. Evidently, these results are as good as the state-of-the-art discussed previously, with the added merit of handling implicit relations in a relatively effective way.

Other works produced in the last decade include the development of a corpus to identify patterns that are both temporal and causal \cite{bethard2008}, ontology enrichment using causation relations \cite{al2013}, and automatic detection of causal relations in Arabic language \cite{sadek2013}. Some application-specific papers consider the extraction of causal networks from news topics \cite{ackerman2012, ishii2010, ishii2012 }, and finding causality in the specialized domain of environment \cite{arauz2012}. We will not discuss these here due to brevity of space.

\section{Statistical \& Machine Learning Techniques}
\label{mlbased}


The need to make use of a large amount of labelled, domain-and-type-independent, textual data and to extract implicit patterns in text automatically meant that machine learning techniques could potentially do much better than purely linguistic techniques. Thus, beginning in the early 2000s, the paradigm to tackle the problem of automatic causal relation extraction  began shifting to statistics and machine learning. The early studies relied on finding explicitly marked cause-effect pairs in sentence, but with the passage of time, researchers progressively began to account for implicit and ambiguous constructs through careful feature extraction.

\subsection{Capturing Explicit Causalities Only via Fixed Linguistic Patterns \& Simple Cue Phrases}

Girju \cite{girju2003}'s venture into the machine learning paradigm was in fact a relatively straight-forward modification of the work by Girju and Moldovan \cite{girju2002} discussed at length in Section \ref{nonmlbased}, where the semi-supervised pattern validation and ranking procedure, consisting of identification of semantic constraints on each causal pattern and pattern ranking, is replaced by a supervised method using C4.5 decision trees \cite{quinlan2014}. A training corpus of 6000 sentences and a test corpus of 1200 sentences containing each of the 60 simple causative verbs was created using a domain-independent text collection. Using a syntactic parser \cite{charniak2000}, 6523 relations of the form \textit{NP1-Verb-NP2} were found, from which 2101 were causal relations and 4422 were not. These were the positive and negative examples used to train the decision-tree classifier. As features, the constraints on the nouns and verb, which were necessary for a pattern to be a causal relation, were identified. In particular, for each value of \textit{NP1} (and similarly for each \textit{NP2}),  nine noun hierarchies from WordNet were used as semantic features: \textit{entity, psychological feature, abstraction, state, event, act, group, possession} and \textit{phenomenon}.
The training process produced several constraints, which were ranked based on frequency and accuracy. The model obtained a precision and recall of 73.91\% and 88.69\% respectively on the test set. The model was also tested on a state-of-the-art Question Answering system \cite{harabagiu2001} with very promising results (see Figure \ref{girju03}). These numbers indicate that this machine learning variant of Girju and Moldovan's work was much more effective, and most of the errors arose from the simplification assumptions, e.g. disambiguating a relation based on a restricted list of 60 causative verbs only. Clearly, the dataset was also very small, which could be a major contributing factor in the errors. 

\begin{figure}
\centering
\includegraphics[width=84mm, height=57mm]{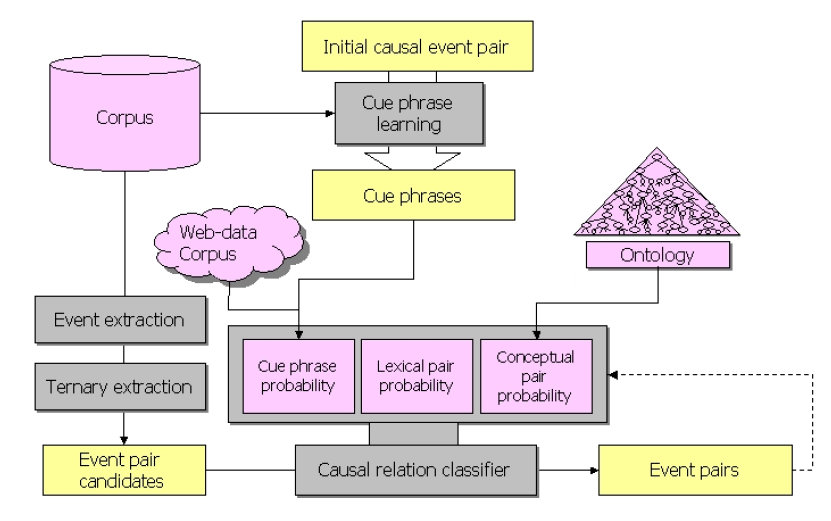}
\caption{Chang and Choi \cite{chang2004}'s causal extraction system.}
\label{changchoi}
\end{figure}

Creating large causally-labelled datasets for supervised machine learning was one way to solve the problem encountered by Girju \cite{girju2003}, but it would be very time and effort consuming. A year later, Chang and Choi \cite{chang2004} came up with a workaround mechanism to learn \textit{cue phrase}\footnote{A cue phrase is a word, phrase or a word pattern that connects one event to another with some relation. The 60 simple causatives found by Girju \cite{girju2003} are examples of causal cue phrases.} and \textit{lexical pair}\footnote{The lexical pair \textit{bacteria-disease} is a causal lexical pair.} probabilities from raw and domain-independent corpus in an unsupervised manner, and use these to extract inter-noun and inter-sentence causalities. This work was inspired from a highly cited work by Marcu and Echihabi \cite{marcu2002} in 2002, which used Naive Bayes classification to differentiate between various inter-sentence semantic relations using lexical pair probabilities. Chang and Choi use a Connexor dependency parser \cite{tapanainen1997} to extract ternary expressions of the form \textit{NP1-CuePhrase-NP2}, where NP1 and NP2 together form the lexical pair. These ternaries are filtered with a set of pre-defined cue phrases based on the 60 causal verbs found in \cite{girju2002, girju2003}. An initial classifier based on cue phrase confidence scores, i.e. the probability of the causal class given a cue phrase, is used to extract highly ranked ternaries, which are treated as the initial causality annotated training set for the Naive Bayes classifier. The Naive Bayes classifier's task is to classify each ternary expression $t_i$ as either causal ($c_1$) or non-causal ($c_0$) using the expression below.
$$ c^* = \argmax_{c_j} P(c_j|t_i) = \argmax_{c_j} \frac{P(c_j)P(t_i|c_j)}{P(t_i}$$
Using the cue phrase $CP_{t_i}$ and the lexical pair $LP_{t_i}$ as the independent features for each training sample $t_i$, 
$$ P(t_i|c_j) = P(CP_{t_i}|c_j) \prod_{k=1}^{|t_i|} P(LP_{k}|c_j),$$
where $P(CP_{t_i}|c_j)$ and  $P(LP_{k}|c_j)$ are the cue phrase and lexical pair probabilities defined earlier. These probabilities and other parameters are learned and improved using an iterative Expectation-Maximization procedure, instead of relying on a large annotated data set. Using an unlabelled domain-independent raw corpus consisting of 5 million sentences as the training set, and two manually annotated test sets\footnote{The authors failed to mention the size of the test sets, but clarified that one set is domain-independent, while the other comes from the medical domain. }, the best results achieved by this model had a precision of 81.29\%. The authors demonstrated how the techniques can be extended to detect inter-sentence causalities as well. An overview of their system's architecture is shown in Figure \ref{changchoi}.

Chang and Choi's work was an effective demonstration of how to extract causal relations based on statistical models using a very large unlabelled data set, but their evaluation results and error analysis left a little something to be desired. They did not quote the model's recall or F-score, indicating that they focused on improving the model precision only, which is not always a good idea.  They also did not justify the use of the Naive Bayes classifier, which over-simplifies the relationship between different features.  Furthermore, the ternary causation pattern used was the same as previous works; so were the cue phrases (simple causative verbs only).  

In 2008, Blanco \textit{et al.}\cite{blanco2008} followed up with a work that considered explicit causal relations expressed by the pattern \textit{VerbPhrase-relator-Cause}, where \textit{relator $\in$ \{because, since, after, as\}}. They used semantically annotated SemCor 2.1 corpus for training a 
C4.5 decision tree binary classifier with bagging. They used seven features for learning: the type of relator, its modifiers, the semantic classes of the verbs, the causal potentials of the verbs, and the tense of the verbs. They obtained an average F-measure of 0.905, in comparison to Girju \cite{girju2003}'s 0.8 and Chang and Choi \cite{chang2004}'s 0.81. Most of the errors were caused by the sentences containing the relators \textit{or} and \textit{after}. This study also made several simplifying assumptions, e.g. considering a single linguistic pattern only which contained marked and explicit causal indicators. Thus, this work could be termed incremental at best.

The year 1998 saw the birth of a forum called SensEval (renamed to SemEval in 2007) for the evaluation of approaches to the semantic analysis tasks in NLP. In SemEval 2007, Girju \text{et al.} \cite{girju2009} proposed Task 4, the classification of seven frequently occuring semantic relations between nouns or noun phrases: Cause-Effect, Instrument-Agency, Product-Producer, Origin-Entity, Theme-Tool, Part-Whole and Content-Container. The best system among fifteen participants in 2007 was by a UIUC team \cite{girju2010}. They used a combination of lexical, syntactic and semantic features over a training set that contained explicit and market causal relations only, and used Support Vector Machines (SVMs) to correctly classify the cause-effect relations with F-measure of 0.82 and accuracy of 0.775. SemEval 2010 task 8 was similar, and Pal \textit{et al.} \cite{pal2010} provided a solution using separate rule-based features for each type of relations based on the verbs and prepositions present between each nominal pair, in conjunction with WordNet bsaed semantic features, to train a Conditional Random Field framework, obtaining an F-score of 0.5216. The best results for SemEval 2010 task 8 were achieved by Rink \textit{et al.} \cite{rink2010utd},obtaining a precision and recall of 0.89. Pakray and Gelbukh \cite{Pakray2014} used WordNet and POS-tagging based dependency relations to train decision trees on the SemEval 2010 datasets, and achieved an F-score of 0.858. See \cite{hendrickx2009} for detailed evaluation of several other systems for this task. Note that the problem of multi-way classification of semantic relations is different from the problem of causal relation extraction. However, one could possibly use a semantic relations classifier for disambiguation during causal relation extraction. To the best of our knowledge, no known system employs this strategy. 

In 2010, Sil \textit{et al.} \cite{sil2010} approached the problem of causality extraction from the perspective of extracting common-sense knowledge from text and, in particular, mining pre-conditions and post-conditions of actions and events from Web text. Thus, in an implicit way, their system, PREPOST, extracts the causal relations an action has with its preconditions and its postconditions. For any action word $A$, Sil \textit{et al.}'s PREPOST uses a search engine to collect a large set of documents $D_A$ that contain $A$ in the pattern \textit{`is/are/was/were A-ing'}. From $D_A$, a list of possible pre- and post-condition words is created by computing the pointwise mutual information (PMI\footnote{PMI is a statistic to measure the degree of association between terms.}) between $A$ and every word in $D_A$ and taking the top 500 words as the candidate pre- and post-conditions. The PMI score encodes the degree of association between the action word and every candidate word. Using this feature, the classifier can determine that `liquid' is a more likely post-condition for `melt' than `gas', because it is more highly correlated with the former in text. Furthermore, to seave out highly correlated words that are not pre- or post-conditions of the action, PREPOST determines the set of feature words that characterize the relationship between the candidate word and the action word, by computing their 3-way PMI with each feature word. This set of feature words is determined by taking the 3000 most frequently occurring words from the entire corpus, computing $\chi^2$ between each candidate feature word and the labeled pre- and post-conditions, and keeping the top 161 feature words corresponding to high values of $\chi^2$.  Several other commonly used semantic feature are extracted from the text annotated with a semantic role labeling (SRL) system \cite{huang2010}. The learning is done using SVMs with  the radial basis function (RBF) kernel. Using 5 action words for training and 35 action words for testing, PREPOST achieved a precision and recall of 0.943 and 0.854 respectively for pre-conditions, with similar values for post-conditions. 

PREPOST only uses a small set of labeled data for learning and generalizes remarkably well to unseen actions and events with high precision and recall. Thus, it is an effective model of how the world changes over time. It also captures pre-conditions relationships, which are not directly captured by any of the previously discussed techniques. Furthermore, techniques mentioned so far cannot distinguished between event-event relationships (rain causes flooding) and event-state relationships (rain causes wet grass). PREPOST focuses on the latter type of relations only. The PMI statistic has also been used by Gordon \textit{et al.} \cite{gordon2011} to extract causal relations between everyday events from personal stories published on the web.

\subsection{Moving Towards Effective Extraction of implicit and Ambiguous Causalities}

So far, no one judiciously tackled the problem of extracting implicit causal relations. In fact, all the proposed statistical and machine learning systems, including the ones mentioned previously, focused on handling disambiguation by carefully choosing the features of the learning algorithm.  Bethard and Martin \cite{bethard2008learning} approached the problem of implicit causal relation detection by building a system that answered the following question: given two events occurring in the same sentence, can one event be considered as the cause of the other? In doing so, they mainly wanted to distinguish between causal and temporal relations, a task not addressed by any previous research. They used surface features, WordNet hypernym\footnote{a word with a broad meaning that more specific words fall under; a superordinate. For example, color is a hypernym of red.} and lexical files, syntactic paths and a score feature for the events based on web counts. Using Google N-gram corpus and SVMs, they achieved an F-score of 0.49 and 0.524 for temporal and causal relations respectively. 

\begin{figure}
\centering
\includegraphics[width=90mm, height=45mm]{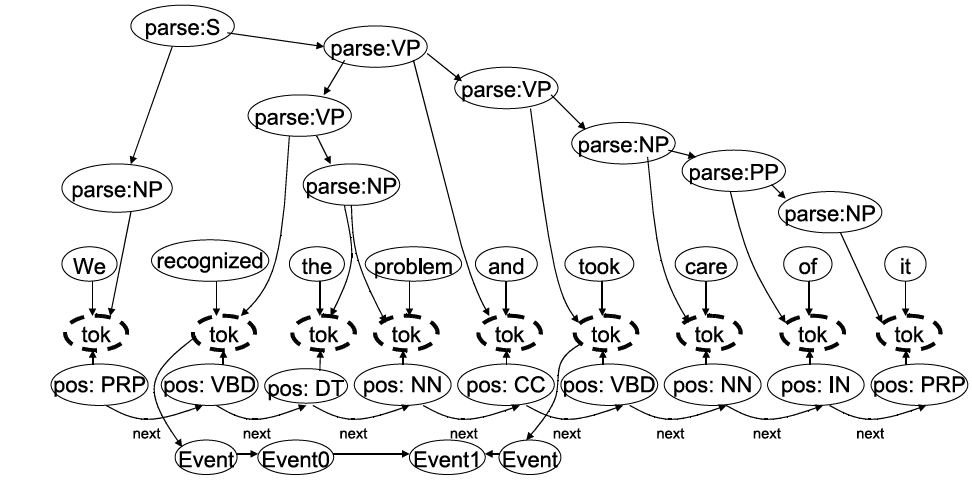}
\caption{An example text graph encoding syntactic information \cite{rink2010}.}
\label{rink10}
\end{figure}

To handle implicit causal relations, Rink \textit{et al.} \cite{rink2010} proposed a novel graphical framework  where the nodes are made up of tokens in a sentence. These graphs can capture lexical and syntactic structure of a sentence as well as semantic relationships such as WSD through special edges and nodes, which allows for pattern matching at a high hypernym level as well down to the most specific sense level (see Figure \ref{rink10}). The task is to extract subgraphs, using a popular Frequent Subgraph Mining algorithm \cite{yan2002}, that represent causal relations. The actual pattern matching within the graph is treated as a constraint satisfaction problem. Different types of graph patterns, e.g.  POS, stems, surface words, hypernym chains and semantic links between verbs, are used as features to train an SVM classifier over the Google N-gram corpus, with the best F1-score of 0.579. This was better than the results obtained by Bethard and Martin, signifying that the graph based method may have significant merits due to its ability to capture dependencies between features, and thus some contextual information as well. Furthermore, this approach also has the advantage that feature combinations and textual structure are automatically discovered rather than manually selected. At the same time, ambiguous verbs like \textit{investigating, said, seeing, saying}, which may or may not be causal depending on the context, contributed to the majority of the errors. 

Sorgente \textit{et al.} \cite{sorgente2013} and Yang and Mao \cite{yang2014} have recently developed sophisticated techniques for the problem at hand. Sorgente \textit{et al.} did pattern matching based on simple causative verbs as well as phrasal verbs (e.g. result in), noun+preposition (e.g. cause of), passive causative verbs (e.g. triggered by), and special single prepositions  (e.g. from, after). Most of the features for each of these constructs are obtained automatically from the relations present in the dependency tree of each sentence \cite{de2006, de2008}. Using a Bayesian classifier and Laplace smoothing on the SemEval 2010 task 8 data set, the best F-score obtained was 0.64. Yang and Mao's MLRE, a multi-level relations extractor, can detect causal relations between clauses, as opposed to nouns or noun phrases only. MLRE is based on the linguistic knowledge of the dependency grammar and constituent grammar, without imposing any restriction on causal relation patterns. In addition to WordNet, two previously unused lexical semantic resources are used for feature acquisition in this paper: VerbNet \cite{schuler2005} and FrameNet \cite{baker1998}. The drawback of the complexity of this approach is that the dataset has to be manually labelled and annotated, which is why it is relatively small and may lead to overfitting. Using an SVM with the RBF kernel and restricted boosting\footnote{Restricted Boosting is a modified version of AdaBoost. }, MLRE achieved an F-score of 0.66. Experiments with MLRE revealed that causal relations indicated by verbs preferred semantic features to syntactic features, whereas the syntactic features made more contribution to relations indicated by prepositions. The rich features based on expert knowledge constitute a novel approach to the problem of implicit causality extraction. 

Finally, due to brevity of space, we briefly mention some recent publications on the applications of causality extraction.\\
- Radinsky \textit{et al.} \cite{radinsky2012}'s PUNDIT uses statistical inferencing and hierarchical clustering to perform \textit{causal prediction}, the process of observing an event and reasoning about future events caused by it. \\
- Do \textit{et al.} \cite{do2011} identify causal relations between events using novel statistical metrics.\\
- Sun \textit{et al.} \cite{sun2007} detect causality between search query pairs in temporal query logs, in an effort to improve the performance of search engines by predicting queries to be made in the future.\\
- Bui \textit{et al.} \cite{bui2010} retrieve causalities between drugs and virus mutations from medical literature using logistic regression.\\
- Krishnan \textit{et al.} \cite{krishnan2014} use Conditional Random Fields to extract causal relations between interventions and disorders from geriatric literature.\\
- Zhao \textit{et al.} \cite{zhao2016} propose a new feature called `causal connectives', obtained by computing the similarity of the syntactic dependency structure of sentences. They also propose a new Restricted Hidden Naive Bayes learning algorithm to cope with the interactions among features, especially those between causal connectives and the lexico-syntactic patterns.

\section{Comparing the Two Paradigms}
\label{analysis}

We have drawn several comparisons between the two paradigms of ML and non-ML techniques.\\
- The early non-ML works suffered from the need to annotate/label the texts manually, and hand-code the linguistic, syntactic and lexical features of interest. This levied size-related constraints on the data sets. In contrast, the early ML research had at their disposal several automatic syntactic and dependency parsers (e.g. \cite{tapanainen1997, de2008}). With the advent of crowdsourcing platforms like Amazon Mechanical Turk, obtaining large labeled datasets should theoretically not be much of a constraint anymore. \\
- The non-ML works predominantly use linguistic and syntactic pattern matching, where the patterns and templates are constructed manually. In comparison, the ML literature focuses on finding these patterns automatically using a small set of seed patterns.\\
- The non-ML techniques were mostly tested on a few different sets of texts, where each set consisted of domain dependent paragraphs/sentences. The ML systems, on the other hand, could be tested on texts containing sentences from various domains.\\
- Each non-ML technique is meant to work only for a specific type of text (episodic, narrative, expository, etc), and may not generalize well to other types. In contrast, the ML techniques are inherently capable of capturing these generalizations, provided that the feature set is appropriate. \\
- In the non-ML papers, causal ambiguities are predominantly captured through manual or semi-supervised semantic ranking measures (e.g. \cite{girju2002}). These ambiguities can be captured automatically  by ML algorithms through class-specific probabilities.\\
- A much higher number of ML techniques consider the extraction of implicit causal relations (i.e. words like `prevent' that are not synonymous with `cause'). This problem was largely ignored by the non-ML works.\\
- Semantic lexicons like WordNet, VerbNet and FrameNet have proven to be extremely useful for both paradigms. While most works focus on WordNet, a few very recent papers attempt to include VerbNet and FrameNet based features with some success. Other broad-coverage corpora like Wikipedia and DBPedia have also been used extensively to create knowledge-bases and ontologies for training.\\
- There is an overwhelming focus on developing domain-independent causality extractors in both paradigms, where broad-coverage corpora like Wikipedia are used for training. However, these systems may not work well on highly specialized domains. Moreover, annotated data for these specialized domains may not be available in abundance, resulting in poor training and generalization.\\
- With a couple of exceptions, none of the papers provide empirical comparisons with existing techniques. This is surprising, and may be due to a general lack of standardized datasets. Thus it is a relatively fruitless exercise to compare the precision, recall or accuracy of one technique with those of another. 

\section{Conclusion \& Future Work}
\label{conc}

In this paper, we have presented a comprehensive survey of research literature, spanning the last three decades, on the automatic extraction of causal relations from the two paradigms of non-statistical and statistical techniques. In addition, we provide an in-depth discussion on at least five highly-cited research studies from both paradigms. We observe that causal relation extraction is one of the most difficult NLP problems primarily due to the presence of implicit linguistic constructs that may or may not be causal, depending on the textual context. Data sparsity is another issue for highly specialized domains where expert annotation is required, making it hard to use minimally supervised techniques for this problem. By using sophisticated feature construction techniques, lexico-semantic resources like WordNet and powerful learning algorithms like SVMs, the machine learning community has been able to achieve reasonably good results in the recent years. However, there is still a long way to go. Some possible avenues for future work are listed below.  \\
- One could build a basic taxonomy of all the techniques mentioned in this survey. We could not include it in this paper due to brevity of time, and plan to do this in the near future.\\
- An empirical comparison of the existing systems is required on standardized data sets.\\ 
- The problem could be modeled using Deep Neural Networks. Their powerful feature abstraction capabilities could effectively capture implicit and ambiguous causal relations, which contribute to most of the errors in the existing systems. Recursive Neural Networks would be well-suited for this task. \\
- Comparing the performance of multiple classifiers (or using ensemble learning) on some of the recent ML-based techniques may make for a good graduate course project.\\
- Combining a generic semantic relations classifier (e.g. SemEval 2010 Task 8) with any existing causal relation extraction system would be a worthwhile attempt at addressing the disambiguation problem.\\

\newpage
\bibliography{refs}

\begin{thebibliography}{10}
\providecommand{\url}[1]{#1}
\csname url@samestyle\endcsname
\providecommand{\newblock}{\relax}
\providecommand{\bibinfo}[2]{#2}
\providecommand{\BIBentrySTDinterwordspacing}{\spaceskip=0pt\relax}
\providecommand{\BIBentryALTinterwordstretchfactor}{4}
\providecommand{\BIBentryALTinterwordspacing}{\spaceskip=\fontdimen2\font plus
\BIBentryALTinterwordstretchfactor\fontdimen3\font minus
  \fontdimen4\font\relax}
\providecommand{\BIBforeignlanguage}[2]{{%
\expandafter\ifx\csname l@#1\endcsname\relax
\typeout{** WARNING: IEEEtran.bst: No hyphenation pattern has been}%
\typeout{** loaded for the language `#1'. Using the pattern for}%
\typeout{** the default language instead.}%
\else
\language=\csname l@#1\endcsname
\fi
#2}}
\providecommand{\BIBdecl}{\relax}
\BIBdecl

\bibitem{girju2003}
R.~Girju, ``Automatic detection of causal relations for question answering,''
  in \emph{Proceedings of the ACL 2003 workshop on Multilingual summarization
  and question answering-Volume 12}.\hskip 1em plus 0.5em minus 0.4em\relax
  Association for Computational Linguistics, 2003, pp. 76--83.

\bibitem{khoo1998}
C.~S. Khoo, J.~Kornfilt, R.~N. Oddy, and S.~H. Myaeng, ``Automatic extraction
  of cause-effect information from newspaper text without knowledge-based
  inferencing,'' \emph{Literary and Linguistic Computing}, vol.~13, no.~4, pp.
  177--186, 1998.

\bibitem{radinsky2012}
K.~Radinsky, S.~Davidovich, and S.~Markovitch, ``Learning causality for news
  events prediction,'' in \emph{Proceedings of the 21st international
  conference on World Wide Web}.\hskip 1em plus 0.5em minus 0.4em\relax ACM,
  2012, pp. 909--918.

\bibitem{silverstein2000}
C.~Silverstein, S.~Brin, R.~Motwani, and J.~Ullman, ``Scalable techniques for
  mining causal structures,'' \emph{Data Mining and Knowledge Discovery},
  vol.~4, no. 2-3, pp. 163--192, 2000.

\bibitem{riaz2010}
M.~Riaz and R.~Girju, ``Another look at causality: Discovering
  scenario-specific contingency relationships with no supervision,'' in
  \emph{Semantic Computing (ICSC), 2010 IEEE Fourth International Conference
  on}.\hskip 1em plus 0.5em minus 0.4em\relax IEEE, 2010, pp. 361--368.

\bibitem{hashimoto2014}
C.~Hashimoto, K.~Torisawa, J.~Kloetzer, M.~Sano, I.~Varga, J.-H. Oh, and
  Y.~Kidawara, ``Toward future scenario generation: Extracting event causality
  exploiting semantic relation, context, and association features.'' in
  \emph{ACL (1)}, 2014, pp. 987--997.

\bibitem{ackerman2012}
E.~J.~M. Ackerman, ``Extracting a causal network of news topics,'' in \emph{On
  the Move to Meaningful Internet Systems: OTM 2012 Workshops}.\hskip 1em plus
  0.5em minus 0.4em\relax Springer, 2012, pp. 33--42.

\bibitem{chan2005}
K.~Chan and W.~Lam, ``Extracting causation knowledge from natural language
  texts,'' \emph{International Journal of Intelligent Systems}, vol.~20, no.~3,
  pp. 327--358, 2005.

\bibitem{khoo2000}
C.~S. Khoo, S.~Chan, and Y.~Niu, ``Extracting causal knowledge from a medical
  database using graphical patterns,'' in \emph{Proceedings of the 38th Annual
  Meeting on Association for Computational Linguistics}.\hskip 1em plus 0.5em
  minus 0.4em\relax Association for Computational Linguistics, 2000, pp.
  336--343.

\bibitem{sachs2005}
K.~Sachs, O.~Perez, D.~Pe'er, D.~A. Lauffenburger, and G.~P. Nolan, ``Causal
  protein-signaling networks derived from multiparameter single-cell data,''
  \emph{Science}, vol. 308, no. 5721, pp. 523--529, 2005.

\bibitem{arauz2012}
P.~L. Ara{\'u}z and P.~Faber, ``Causality in the specialized domain of the
  environment,'' in \emph{Semantic Relations-II. Enhancing Resources and
  Applications Workshop Programme}.\hskip 1em plus 0.5em minus 0.4em\relax
  Citeseer, 2012, p.~10.

\bibitem{wolff2003}
P.~Wolff and G.~Song, ``Models of causation and the semantics of causal
  verbs,'' \emph{Cognitive Psychology}, vol.~47, no.~3, pp. 276--332, 2003.

\bibitem{wolff2007}
P.~Wolff, ``Representing causation.'' \emph{Journal of experimental psychology:
  General}, vol. 136, no.~1, p.~82, 2007.

\bibitem{talmy2000}
L.~Talmy, \emph{Toward a cognitive semantics}.\hskip 1em plus 0.5em minus
  0.4em\relax MIT press, 2000, vol.~1.

\bibitem{hobbs2005}
J.~R. Hobbs, ``Toward a useful concept of causality for lexical semantics,''
  \emph{Journal of Semantics}, vol.~22, no.~2, pp. 181--209, 2005.

\bibitem{white1990}
P.~A. White, ``Ideas about causation in philosophy and psychology.''
  \emph{Psychological bulletin}, vol. 108, no.~1, p.~3, 1990.

\bibitem{altenberg1984}
B.~Altenberg, ``Causal linking in spoken and written english,'' \emph{Studia
  linguistica}, vol.~38, no.~1, pp. 20--69, 1984.

\bibitem{thomson1987}
J.~J. Thomson, ``Verbs of action,'' \emph{Synthese}, vol.~72, no.~1, pp.
  103--122, 1987.

\bibitem{simpson1983}
J.~Simpson, ``Resultatives,'' 1983.

\bibitem{cresswell1981}
M.~J. Cresswell, ``Adverbs of causation,'' in \emph{Adverbial
  Modification}.\hskip 1em plus 0.5em minus 0.4em\relax Springer, 1981, pp.
  173--192.

\bibitem{nastase2004}
V.~A. Nastase, \emph{Semantic relations across syntactic levels}.\hskip 1em
  plus 0.5em minus 0.4em\relax University of Ottawa, 2004.

\bibitem{blanco2008}
E.~Blanco, N.~Castell, and D.~I. Moldovan, ``Causal relation extraction.'' in
  \emph{LREC}, 2008.

\bibitem{sadek2013}
J.~Sadek, ``Automatic detection of arabic causal relations,'' in \emph{Natural
  Language Processing and Information Systems}.\hskip 1em plus 0.5em minus
  0.4em\relax Springer, 2013, pp. 400--403.

\bibitem{garcia1997}
D.~Garcia, ``Coatis, an nlp system to locate expressions of actions connected
  by causality links,'' in \emph{Knowledge Acquisition, Modeling and
  Management}.\hskip 1em plus 0.5em minus 0.4em\relax Springer, 1997, pp.
  347--352.

\bibitem{selfridge1989}
M.~Selfridge, ``Toward a natural language-based causal model acquisition
  system,'' \emph{Applied Artificial Intelligence an International Journal},
  vol.~3, no. 2-3, pp. 191--212, 1989.

\bibitem{kaplan1991}
R.~M. Kaplan and G.~Berry-Rogghe, ``Knowledge-based acquisition of causal
  relationships in text,'' \emph{Knowledge Acquisition}, vol.~3, no.~3, pp.
  317--337, 1991.

\bibitem{doyle1984}
R.~J. Doyle, ``Hypothesizing and refining causal models,'' DTIC Document, Tech.
  Rep., 1984.

\bibitem{grishman1990}
R.~Grishman, ``Domain modeling for language analysis,'' \emph{Linguistic
  approaches to artificial intelligence}, pp. 41--58, 1990.

\bibitem{kontos1991}
J.~Kontos and M.~Sidiropoulou, ``On the acquisition of causal knowledge from
  scientific texts with attribute grammars,'' \emph{International Journal of
  Applied Expert Systems}, vol.~4, no.~1, pp. 31--48, 1991.

\bibitem{khoo1999}
C.~S. Khoo, S.~Chan, Y.~Niu, and A.~Ang, ``A method for extracting causal
  knowledge from textual databases,'' \emph{Singapore journal of library \&
  information management}, vol.~28, pp. 48--63, 1999.

\bibitem{khoo1995}
C.~S. Khoo, ``Automatic identification of causal relations in text and their
  use for improving precision in information retrieval,'' Ph.D. dissertation,
  1995.

\bibitem{finn1998}
R.~Finn, ``Program uncovers hidden connections in the literature,'' \emph{The
  Scientist}, vol.~12, no.~10, pp. 12--13, 1998.

\bibitem{girju2002}
R.~Girju and D.~Moldovan, ``Text mining for causal relations.'' in \emph{FLAIRS
  Conference}, 2002, pp. 360--364.

\bibitem{nedjalkov1973}
V.~Nedjalkov and G.~Silnickij, ``The topology of causative constructions,''
  \emph{Folia linguistica}, vol.~6, pp. 273--290, 1973.

\bibitem{hearst1998}
M.~A. Hearst, ``Automated discovery of wordnet relations,'' \emph{WordNet: an
  electronic lexical database}, pp. 131--153, 1998.

\bibitem{miller1995}
G.~A. Miller, ``Wordnet: a lexical database for english,'' \emph{Communications
  of the ACM}, vol.~38, no.~11, pp. 39--41, 1995.

\bibitem{low2001}
B.-T. Low, K.~Chan, L.-L. Choi, M.-Y. Chin, and S.-L. Lay, ``Semantic
  expectation-based causation knowledge extraction: A study on hong kong stock
  movement analysis,'' in \emph{Advances in Knowledge Discovery and Data
  Mining}.\hskip 1em plus 0.5em minus 0.4em\relax Springer, 2001, pp. 114--123.

\bibitem{chan2002}
K.~Chan, B.-T. Low, W.~Lam, and K.-P. Lam, \emph{Extracting causation knowledge
  from natural language texts}.\hskip 1em plus 0.5em minus 0.4em\relax
  Springer, 2002.

\bibitem{ittoo2011}
A.~Ittoo and G.~Bouma, ``Extracting explicit and implicit causal relations from
  sparse, domain-specific texts,'' in \emph{Natural Language Processing and
  Information Systems}.\hskip 1em plus 0.5em minus 0.4em\relax Springer, 2011,
  pp. 52--63.

\bibitem{bethard2008}
S.~Bethard, W.~J. Corvey, S.~Klingenstein, and J.~H. Martin, ``Building a
  corpus of temporal-causal structure.'' in \emph{LREC}, 2008.

\bibitem{al2013}
H.~Al~Hashimy, A.~Saleh, and N.~Kulathuramaiyer, ``Ontology enrichment with
  causation relations,'' in \emph{Systems, Process \& Control (ICSPC), 2013
  IEEE Conference on}.\hskip 1em plus 0.5em minus 0.4em\relax IEEE, 2013, pp.
  186--192.

\bibitem{ishii2010}
H.~Ishii, Q.~Ma, and M.~Yoshikawa, ``Causal network construction to support
  understanding of news,'' in \emph{System Sciences (HICSS), 2010 43rd Hawaii
  International Conference on}.\hskip 1em plus 0.5em minus 0.4em\relax IEEE,
  2010, pp. 1--10.

\bibitem{ishii2012}
------, ``Incremental construction of causal network from news articles,''
  \emph{Information and Media Technologies}, vol.~7, no.~1, pp. 110--118, 2012.

\bibitem{quinlan2014}
J.~R. Quinlan, \emph{C4. 5: programs for machine learning}.\hskip 1em plus
  0.5em minus 0.4em\relax Elsevier, 2014.

\bibitem{charniak2000}
E.~Charniak, ``A maximum-entropy-inspired parser,'' in \emph{Proceedings of the
  1st North American chapter of the Association for Computational Linguistics
  conference}.\hskip 1em plus 0.5em minus 0.4em\relax Association for
  Computational Linguistics, 2000, pp. 132--139.

\bibitem{harabagiu2001}
S.~M. Harabagiu, D.~I. Moldovan, M.~Pa{\c{s}}ca, M.~Surdeanu, R.~Mihalcea,
  C.~R. G{\^\i}rju, V.~Rus, F.~L{\u{a}}c{\u{a}}tu{\c{s}}u, P.~Mor{\u{a}}rescu,
  and R.~Bunescu, ``Answering complex, list and context questions with lcc's
  question-answering server,'' 2001.

\bibitem{chang2004}
D.-S. Chang and K.-S. Choi, ``Causal relation extraction using cue phrase and
  lexical pair probabilities,'' in \emph{Natural Language Processing--IJCNLP
  2004}.\hskip 1em plus 0.5em minus 0.4em\relax Springer, 2004, pp. 61--70.

\bibitem{marcu2002}
D.~Marcu and A.~Echihabi, ``An unsupervised approach to recognizing discourse
  relations,'' in \emph{Proceedings of the 40th Annual Meeting on Association
  for Computational Linguistics}.\hskip 1em plus 0.5em minus 0.4em\relax
  Association for Computational Linguistics, 2002, pp. 368--375.

\bibitem{tapanainen1997}
P.~Tapanainen and T.~J{\"a}rvinen, ``A non-projective dependency parser,'' in
  \emph{Proceedings of the fifth conference on Applied natural language
  processing}.\hskip 1em plus 0.5em minus 0.4em\relax Association for
  Computational Linguistics, 1997, pp. 64--71.

\bibitem{girju2009}
R.~Girju, P.~Nakov, V.~Nastase, S.~Szpakowicz, P.~Turney, and D.~Yuret,
  ``Classification of semantic relations between nominals,'' \emph{Language
  Resources and Evaluation}, vol.~43, no.~2, pp. 105--121, 2009.

\bibitem{girju2010}
R.~Girju, B.~Beamer, A.~Rozovskaya, A.~Fister, and S.~Bhat, ``A knowledge-rich
  approach to identifying semantic relations between nominals,''
  \emph{Information processing \& management}, vol.~46, no.~5, pp. 589--610,
  2010.

\bibitem{pal2010}
S.~Pal, P.~Pakray, D.~Das, and S.~Bandyopadhyay, ``Ju: a supervised approach to
  identify semantic relations from paired nominals,'' in \emph{Proceedings of
  the 5th International Workshop on Semantic Evaluation}.\hskip 1em plus 0.5em
  minus 0.4em\relax Association for Computational Linguistics, 2010, pp.
  206--209.

\bibitem{rink2010utd}
B.~Rink and S.~Harabagiu, ``Utd: Classifying semantic relations by combining
  lexical and semantic resources,'' in \emph{Proceedings of the 5th
  International Workshop on Semantic Evaluation}.\hskip 1em plus 0.5em minus
  0.4em\relax Association for Computational Linguistics, 2010, pp. 256--259.

\bibitem{Pakray2014}
P.~Pakray and A.~Gelbukh, ``An open-domain cause-effect relation detection from
  paired nominals,'' in \emph{Nature-Inspired Computation and Machine
  Learning}.\hskip 1em plus 0.5em minus 0.4em\relax Springer, 2014, pp.
  263--271.

\bibitem{hendrickx2009}
I.~Hendrickx, S.~N. Kim, Z.~Kozareva, P.~Nakov, D.~{\'O}~S{\'e}aghdha,
  S.~Pad{\'o}, M.~Pennacchiotti, L.~Romano, and S.~Szpakowicz, ``Semeval-2010
  task 8: Multi-way classification of semantic relations between pairs of
  nominals,'' in \emph{Proceedings of the Workshop on Semantic Evaluations:
  Recent Achievements and Future Directions}.\hskip 1em plus 0.5em minus
  0.4em\relax Association for Computational Linguistics, 2009, pp. 94--99.

\bibitem{sil2010}
A.~Sil, F.~Huang, and A.~Yates, ``Extracting action and event semantics from
  web text.'' in \emph{AAAI Fall Symposium: Commonsense Knowledge}, 2010.

\bibitem{huang2010}
F.~Huang and A.~Yates, ``Open-domain semantic role labeling by modeling word
  spans,'' in \emph{Proceedings of the 48th Annual Meeting of the Association
  for Computational Linguistics}.\hskip 1em plus 0.5em minus 0.4em\relax
  Association for Computational Linguistics, 2010, pp. 968--978.

\bibitem{gordon2011}
A.~S. Gordon, C.~A. Bejan, and K.~Sagae, ``Commonsense causal reasoning using
  millions of personal stories.'' in \emph{AAAI}, 2011.

\bibitem{bethard2008learning}
S.~Bethard and J.~H. Martin, ``Learning semantic links from a corpus of
  parallel temporal and causal relations,'' in \emph{Proceedings of the 46th
  Annual Meeting of the Association for Computational Linguistics on Human
  Language Technologies: Short Papers}.\hskip 1em plus 0.5em minus 0.4em\relax
  Association for Computational Linguistics, 2008, pp. 177--180.

\bibitem{rink2010}
B.~Rink, C.~A. Bejan, and S.~M. Harabagiu, ``Learning textual graph patterns to
  detect causal event relations.'' in \emph{FLAIRS Conference}, 2010.

\bibitem{yan2002}
X.~Yan and J.~Han, ``gspan: Graph-based substructure pattern mining,'' in
  \emph{Data Mining, 2002. ICDM 2003. Proceedings. 2002 IEEE International
  Conference on}.\hskip 1em plus 0.5em minus 0.4em\relax IEEE, 2002, pp.
  721--724.

\bibitem{sorgente2013}
A.~Sorgente, G.~Vettigli, and F.~Mele, ``Automatic extraction of cause-effect
  relations in natural language text.'' \emph{DART@ AI* IA}, vol. 2013, pp.
  37--48, 2013.

\bibitem{yang2014}
X.~Yang and K.~Mao, ``Multi level causal relation identification using extended
  features,'' \emph{Expert Systems with Applications}, vol.~41, no.~16, pp.
  7171--7181, 2014.

\bibitem{de2006}
M.-C. De~Marneffe, B.~MacCartney, C.~D. Manning \emph{et~al.}, ``Generating
  typed dependency parses from phrase structure parses,'' in \emph{Proceedings
  of LREC}, vol.~6, no. 2006, 2006, pp. 449--454.

\bibitem{de2008}
M.-C. De~Marneffe and C.~D. Manning, ``Stanford typed dependencies manual,''
  Technical report, Stanford University, Tech. Rep., 2008.

\bibitem{schuler2005}
K.~K. Schuler, ``Verbnet: A broad-coverage, comprehensive verb lexicon,'' 2005.

\bibitem{baker1998}
C.~F. Baker, C.~J. Fillmore, and J.~B. Lowe, ``The berkeley framenet project,''
  in \emph{Proceedings of the 17th international conference on Computational
  linguistics-Volume 1}.\hskip 1em plus 0.5em minus 0.4em\relax Association for
  Computational Linguistics, 1998, pp. 86--90.

\bibitem{do2011}
Q.~X. Do, Y.~S. Chan, and D.~Roth, ``Minimally supervised event causality
  identification,'' in \emph{Proceedings of the Conference on Empirical Methods
  in Natural Language Processing}.\hskip 1em plus 0.5em minus 0.4em\relax
  Association for Computational Linguistics, 2011, pp. 294--303.

\bibitem{sun2007}
Y.~Sun, K.~Xie, N.~Liu, S.~Yan, B.~Zhang, and Z.~Chen, ``Causal relation of
  queries from temporal logs,'' in \emph{Proceedings of the 16th international
  conference on World Wide Web}.\hskip 1em plus 0.5em minus 0.4em\relax ACM,
  2007, pp. 1141--1142.

\bibitem{bui2010}
Q.-C. Bui, B.~{\'O}. Nuall{\'a}in, C.~A. Boucher, and P.~M. Sloot, ``Extracting
  causal relations on hiv drug resistance from literature,'' \emph{BMC
  bioinformatics}, vol.~11, no.~1, p.~1, 2010.

\bibitem{ittoo2013}
A.~Ittoo and G.~Bouma, ``Minimally-supervised learning of domain-specific
  causal relations using an open-domain corpus as knowledge base,'' \emph{Data
  \& Knowledge Engineering}, vol.~88, pp. 142--163, 2013.

\bibitem{krishnan2014}
A.~Krishnan, J.~Sligh, E.~Tinsley, N.~Crohn, J.~Bandos, H.~Bush, J.~Depasquale,
  and M.~Palakal, ``Causal association mining from geriatric literature,'' in
  \emph{Bioinformatics and Bioengineering (BIBE), 2014 IEEE International
  Conference on}.\hskip 1em plus 0.5em minus 0.4em\relax IEEE, 2014, pp.
  226--230.

\bibitem{zhao2016}
S.~Zhao, T.~Liu, S.~Zhao, Y.~Chen, and J.-Y. Nie, ``Event causality extraction
  based on connectives analysis,'' \emph{Neurocomputing}, vol. 173, pp.
  1943--1950, 2016.

\end{thebibliography}
\bibliographystyle{IEEEtran}
%




\end{document}